\documentclass[9pt,a4paper]{article}
\usepackage{subcaption}
\usepackage{siunitx}
\usepackage{tikz,tkz-tab}
\usepackage{authblk}
\usepackage{verbatim}   	
\usepackage{siunitx,graphicx,overpic,epstopdf,amsmath,amssymb,bm}

\begin{document}

\title{Fast and robust misalignment correction of Fourier ptychographic microscopy}

\author[1,2]{Ao Zhou}
\author[3]{Wei Wang}
\author[1,4,*]{Ni Chen}
\author[5]{Edmund Y. Lam}
\author[4]{Byoungho Lee}
\author[1]{Guohai Situ}

\affil[1]{Shanghai Institute of Optics and Fine Mechanics, Chinese Academy of Sciences, Shanghai 201800, China}
\affil[2]{University of Chinese Academy of Sciences, Beijing 100049, China.}
\affil[3]{Mechanobiology Institute, National University of Singapore, 5A Engineering Drive 1, 117411, Singapore.}
\affil[4]{Department of Electrical and Computer Engineering,
Seoul National University, Seoul 08826, Korea.}
\affil[5]{Department of Electrical and Electronic Engineering, The University of Hong Kong, Hong Kong.}
\affil[*]{Corresponding author: nichen@siom.ac.cn}
   
\maketitle


\begin{abstract}
Fourier ptychographic microscopy~(FPM) is a newly developed computational imaging technique that can provide gigapixel images with both high resolution~(HR) and wide field of view~(FOV). However, the position misalignment of the LED array induces a degradation of the reconstructed image, especially in the regions away from the optical axis. In this paper, we propose a robust and fast method to correct the LED misalignment of the FPM, termed as misalignment correction for the FPM~(mcFPM). Although different regions in the FOV have different sensitivity to the LED misalignment, the experimental results show that the mcFPM is robust with respect to the elimination of each region. Compared with the state-of-the-art methods, the mcFPM is much faster.
\end{abstract}

\section{Introduction}

As we all know, almost all of the conventional microscope has a trade-off between its resolution and FOV. To solve this problem, a new computational imaging technique called FPM has been proposed~\cite{zheng2013wide,zheng2014breakthroughs}. In a typical FPM system, a programmable LED array is used instead of the conventional microscope's light source for providing angularly variant illumination. After capturing a sequence of low resolution~(LR) images under different illumination angles, an iterative phase retrieval process~\cite{fienup1982phase,maiden2009improved,Chen_2014_JOSK} is used to stitch together those LR images in the Fourier space, and then an HR and high space-bandwidth product~(SBP) complex field of the sample can be recovered. Compared with the conventional microscopy, the FPM can achieve HR, wide FOV and quantitative phase imaging~\cite{ou2013quantitative}. Therefore, it has great potential in a variety of applications, such as biomedical medicine ~\cite{williams2014fourier,horstmeyer2015digital,chung2015counting}, characterizing unknown optical aberrations of lenses~\cite{bian2013adaptive,ou2014embedded}.

The FPM shares its roots of ptychography~\cite{rodenburg1992theory,faulkner2004movable,rodenburg2007hard}. In the conventional ptychography, the mechanical scanning in the imaging process makes the position correction of the probe function essential. Similar to the conventional ptychography, the position misalignment of the LED array in the FPM is of great importance. Because the position of each LED determines the wave-vector of the illumination, misalignment of it induces significant errors to the pupil function in the reconstruction process. In the conventional ptychography, a simulated annealing~(SA) algorithm was adopted to correct the position errors of the probe function~\cite{maiden2012annealing}. Similarly, to correct the LED misalignment in the FPM, the SA algorithm has been introduced into each sub-iteration of the reconstruction algorithm. An optimal shift of the pupil function in the Fourier domain can be obtained during this process. However, this method may lead to an algorithmic disorder of the LED array, resulting in the degradation of the reconstruction~\cite{yeh2015experimental}. To avoid this problem, a position correction approach, named as pcFPM~\cite{sun2016efficient}, has been proposed. It is based on the SA algorithm and a non-linear regression technique. In the pcFPM, a global position misalignment model of the LED array was introduced to ensure the corrected LEDs are positionally ordered. Although the pcFPM can effectively eliminate the LED misalignment, the additional process of non-linear regression increases the algorithm complexity and computer load. Actually, during the FPM reconstruction process, it usually requires to divide the captured images into many small segments. However, with these existing methods~\cite{yeh2015experimental,sun2016efficient}, we find that the global shift of the LED array corrected according to the central segment of the FOV may not work for all of the segments. Influenced by the system errors and lens aberration, it seems that different segment requires different global shift of the LED array. Although these methods can correct the global shifts of different segments respectively, they are heavily time consuming.

In this paper, we propose a simple and fast method to correct the global shift of the LED array for the whole FOV. Rather than correcting the shift errors of the pupil function in the Fourier domain, we introduce a global position misalignment model, and then directly correct the global shift of the LED array. The experimental results show the proposed mcFPM is robust and faster than the other methods.

This paper is arranged as the following: In section 2, we will investigate the problem of LED misalignment in the FPM and propose our mcFPM. In section 3, we will present experimental results to verify the effectiveness of the proposed mcFPM. Finally, we will make a conclusion.

\section{Principle}

\subsection{Forward imaging model of the FPM}
\begin{figure}[htbp]
	\centering
	\fbox{\includegraphics[width=0.8\linewidth]{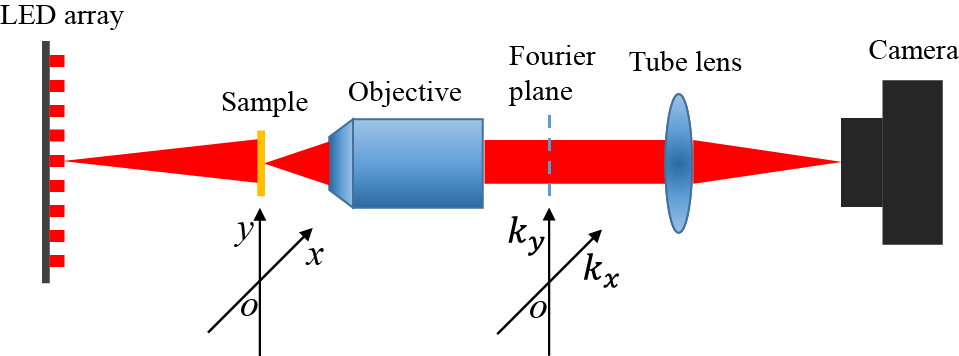}}
	\caption{The imaging process of the FPM.}
	\label{fig1}
\end{figure}

Before introducing the impact of the LED array position misalignment in the FPM, we first introduce the forward imaging model of it. As Fig.~\ref{fig1} shows, in the imaging process, the LEDs are switched on sequentially, and the intensity images of the sample under different illumination angles are captured. Let $A_{object}(x,y)$ represent the complex amplitude of the sample, $A_{output}(x,y)$ represent the output complex amplitude of the sample, and $h(x,y)$ represent the coherent point spread function. When the $m^{th}$ row and $n^{th}$ column LED in the array is on, this process can be modeled as~\cite{zheng2015fourier}:
\begin{equation}
	A_{output}\left(x, y\right) = \left(A_{object}\left(x, y\right) \text{e}^{ik_x^{m,n} x + ik_y^{m,n} y}\right) \otimes h(x,y),
	\label{eq1}
\end{equation}
where $(k_x^{m,n}, k_y^{m,n})$ represents the wave-vector of the parallel light, and $\otimes$ the 2D convolution operator. In the Fourier domain, Eq.~(\ref{eq1}) can be written as
\begin{equation}
	G_{output}\left(k_x, k_y\right) = G_{object}\left( k_x-k_x^{m,n}, k_y-k_y^{m,n}\right) H\left(k_x, k_y\right),
	\label{eq2}
\end{equation}
where $G_{object}(k_x, k_y)$ represents the object spectrum, $G_{output}(k_x, k_y)$ the output spectrum of the microscope, and $H(k_x, k_y)$ the coherent transform function of the microscope. The image captured by the camera thus can be written as
\begin{equation}
	I_{captured}^{m,n}(x, y)\ = \left|\mathcal{F}^{-1}\left\lbrace G_{output}\left(k_x, k_y\right) \right\rbrace \right|^2 
	\label{eq3}
\end{equation}
where $\mathcal{F}^{-1}$ represents the inverse Fourier transform.

With a sequence of LR images captured under different illumination angles, an HR image of the sample can be reconstructed. The recovery process of the FPM follows the strategy of the phase retrieval technique~\cite{fienup1982phase,maiden2009improved}. The algorithm switches between the spatial and Fourier domains. In the spatial domain, the LR intensity measurements are used as the object constraints to ensure the solution convergence. In the Fourier domain, the confined coherent transfer function of the objective lens is imposed as the support constraint. After several iterations, both HR complex field of object and the pupil function will be obtained.

\subsection{Global shift of the LED array in the FPM}

\begin{figure}[htbp]
	\centering
	\fbox{\includegraphics[width=0.8\linewidth]{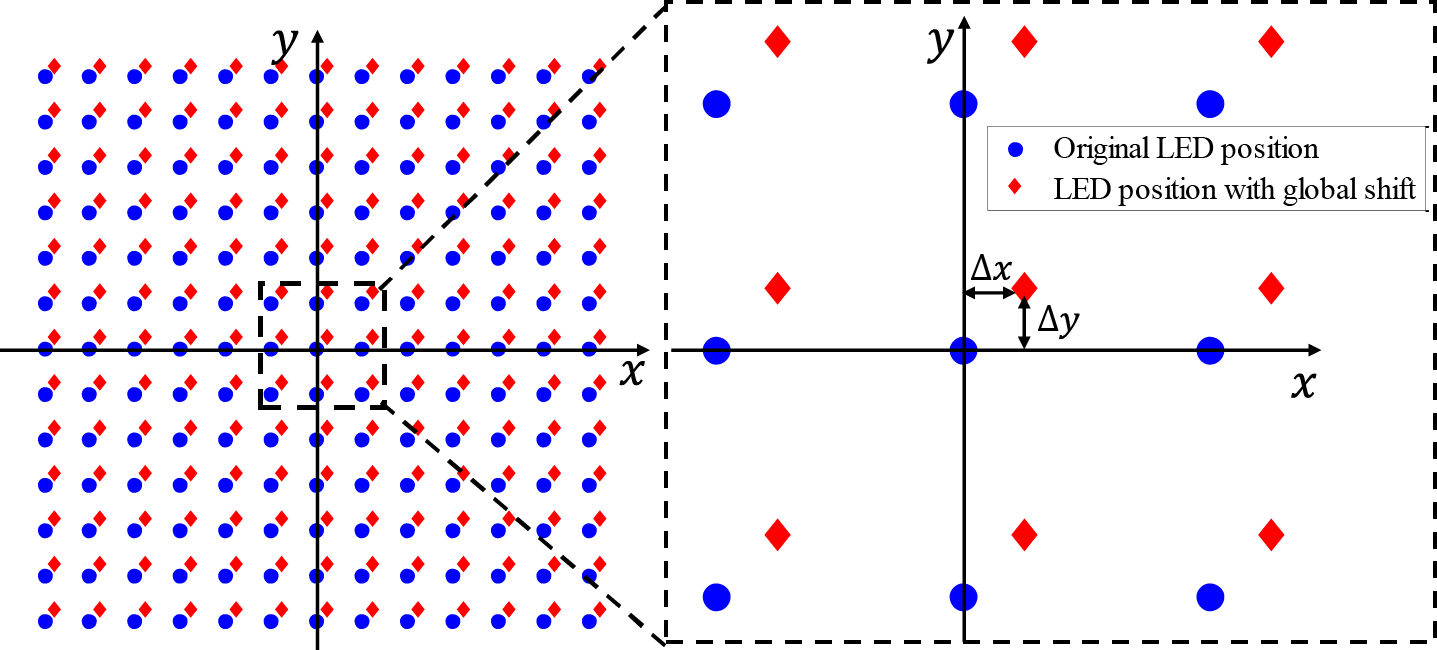}}
	\caption{The global shift model of the LED array in the FPM setup.}
	\label{fig2}
\end{figure}

In the FPM system setup, it requires that the central LED of the LED array is on the optical axis. In the pcFPM~\cite{sun2016efficient}, four global positional factors~(rotation, horizontal and vertical shifts) of the LED array were introduced. However, compared with the other factors, the horizontal shift is hard to eliminate by the hardware. Details will be shown in section 3. In this work, to simplify the position model, a global model with two factors is proposed, as shown in Fig.\ref{fig2}. The two factors that can determine the position of each LED, are marked as $\Delta x$, $\Delta y$. The position of each LED then can be written as
\begin{align}
	&x_{m,n} = m d + \Delta x ,\nonumber \\
	&y_{m,n} = n d + \Delta y ,
	\label{eq4}
\end{align}
where $(x_{m,n}$, $y_{m,n})$ represents the position of the LED at the $m^{th}$ row and $n^{th}$ column, and $d$ the distance between two adjacent LED elements. The incident wave-vector $(k_x^{m,n}, k_y^{m,n})$ for each segment can be written as~\cite{zheng2013wide}
\begin{align}
	&k_x^{m,n} = -\frac{2\pi}{\lambda}\frac{x_o-x_{m,n}}{\sqrt{(x_o-x_{m,n})^2 + (y_o-y_{m,n})^2 + s^2}},\nonumber \\
	&k_y^{m,n} = -\frac{2\pi}{\lambda}\frac{y_o-y_{m,n}}{\sqrt{(x_o-x_{m,n})^2 + (y_o-y_{m,n})^2 + s^2}},
	\label{eq5}
\end{align}
where $(x_o, y_o)$ is the central coordinate of the sample, $\lambda$ is the central wavelength of the LED, and $s$ is the distance between the sample and the LED array.

\begin{figure}[htb]
	\centering
	\captionsetup[subfigure]{justification=centering}
	\begin{subfigure}[b]{0.2\linewidth}
		\centering
		\includegraphics[width=\linewidth]{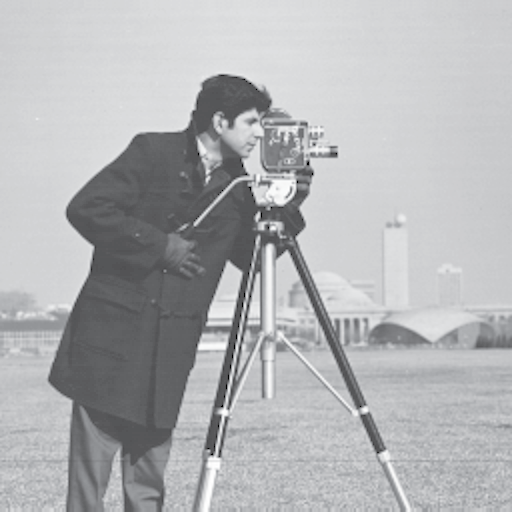}
		\caption{}
	\end{subfigure}
	\begin{subfigure}[b]{0.2\linewidth}
		\centering
		\includegraphics[width=\linewidth]{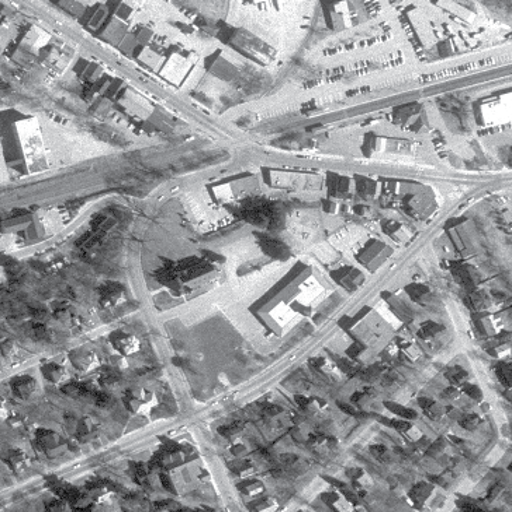}
		\caption{}
	\end{subfigure}	
	\begin{subfigure}[b]{0.2\linewidth}
		\centering
		\includegraphics[width=\linewidth]{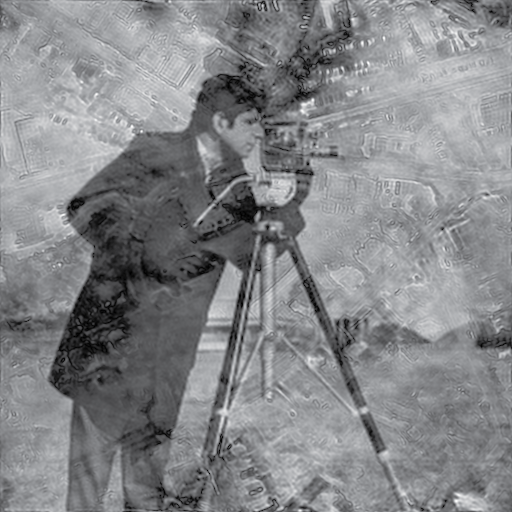}
		\caption{}
	\end{subfigure}
	\begin{subfigure}[b]{0.2\linewidth}
		\centering
		\includegraphics[width=\linewidth]{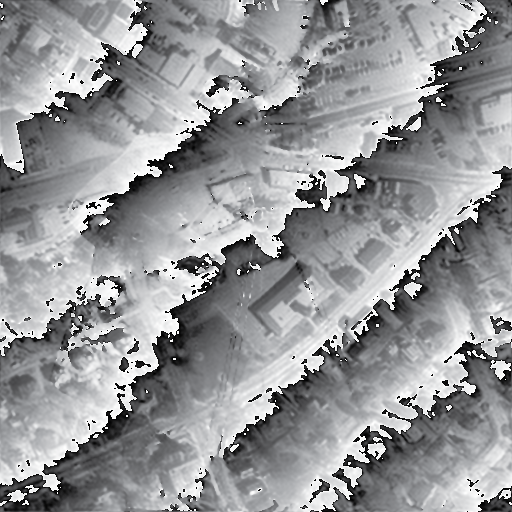}
		\caption{}
	\end{subfigure}
	\caption{\label{fig3} A simulated reconstruction with a global shift of the LED array in the FPM. (a) Amplitude and (b) phase profiles of the object, the reconstructed (c) amplitude and (d) phase profiles with a global shift of LED array }
\end{figure}

From Eq.~(\ref{eq5}), we can find that the global shift of the LED array is a factor that determines the incident wave-vector. Equation~(\ref{eq2}) shows that the error of the incident wave-vector can cause a dislocation  of the object spectrum. In other words, during the FPM reconstruction process, the global shift can induce a shift error to the pupil function in the Fourier domain. Figure~\ref{fig3} is a simulation example. Figure~\ref{fig3}(c) and \ref{fig3}(d) show the reconstructed amplitude and phase profiles of the object with a global shift of the LED array. Compared with the original amplitude and phase profiles in Fig.~\ref{fig3}(a) and \ref{fig3}(b), we can obviously observe the image degradation. This is similar to the reconstructed images in the regions away from the optical axis we observed in the experiments.

\subsection{Proposed LED misalignment correction method} 

\begin{figure}[htb]
	\centering
	\includegraphics[width=.65\linewidth]{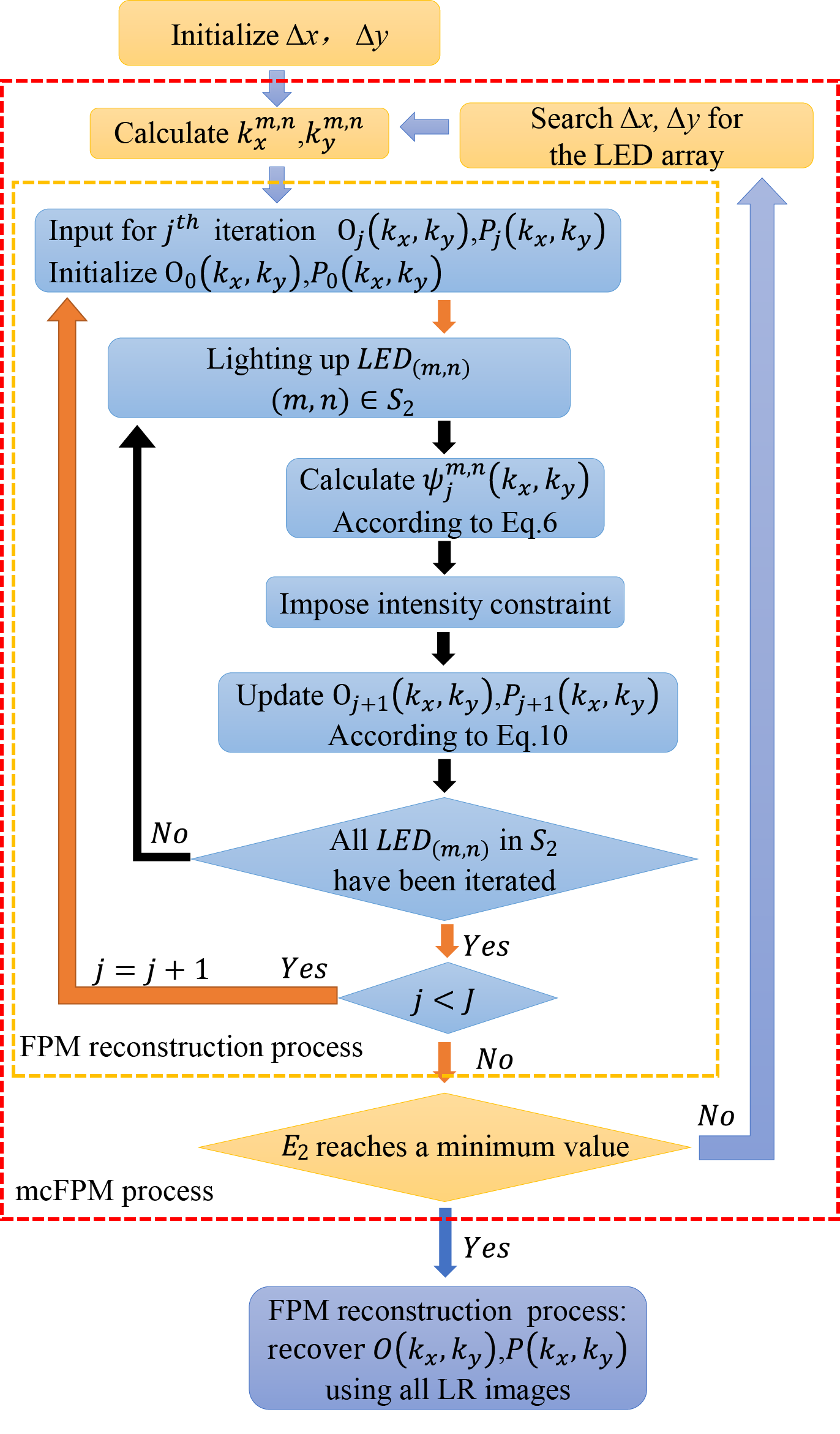}
	\caption{\label{fig4} The flow chart of the mcFPM.}
\end{figure}

The flowchart of the proposed mcFPM is shown in Fig.~\ref{fig4}. Actually, the conventional FPM algorithm is a part of the proposed mcFPM, as the part within the yellow dashed rectangle in Fig.~\ref{fig4}. It is sketched as follows~\cite{zheng2013wide,tian2014multiplexed}:

Step 1. Initialize the Fourier spectrum of the reconstructed HR object $O_j(k_x, k_y)$ and the pupil function $P_j(k_x, k_y)$.

Step 2. Generate an LR image corresponding to the $m^{th}$ row and $n^{th}$ column LED with the incident wave-vector of $(k_x^{m,n}, k_y^{m,n})$ by the below equation
\begin{equation}
\varPsi_j^{m,n}\left(k_x, k_y\right) = O_j\left( k_x-k_x^{m,n}, k_y-k_y^{m,n}\right) P_j\left(k_x, k_y\right),
\label{eq6}
\end{equation}
where $\varPsi_j^{m,n}(k_x, k_y)$ represents the Fourier spectrum of the LR image obtained by illuminating the sample by the $m^{th}$ row and $n^{th}$ column LED. 

Step 3. Impose the intensity constraint with the captured images with
\begin{equation}
\phi_j^{m,n}\left(x, y\right) = \sqrt{\frac{I_{captured}^{m,n}\left(x, y\right)}{\left|\psi_j^{m,n}\left(x, y\right)\right|^2}} \psi_j^{m,n}\left(x, y\right),
\label{eq7}
\end{equation}
where $\phi_j^{m,n}\left(x, y\right)$ and $\psi_j^{m,n}\left(x, y\right)$ are the complex field of the LR images with and without the intensity constraint respectively, and 
\begin{equation}
\psi_j^{m,n}\left(x, y\right) = \mathcal{F}^{-1}\left\lbrace \varPsi_j^{m,n}\left( k_x, k_y\right)\right\rbrace  .
\label{eq8}
\end{equation}

At this time, the updated Fourier spectrum of the LR image is 
\begin{equation}
\varPhi\left(k_x, k_y\right) = \mathcal{F}\left\lbrace\phi_j^{m,n}\left(x, y\right)\right\rbrace,
\label{eq9}
\end{equation}

Step 4. Update the object and the pupil functions with
\begin{align}
&O_{j+1}\left(k_x, k_y\right) = O_{j}\left(k_x, k_y\right) 
+ \frac{\left| P_j\left( k_x+k_x^{m,n}, k_y+k_y^{m,n}\right)\right| P_j^*\left(k_x+k_x^{m,n}, k_y+k_y^{m,n}\right)} {\left| P_j\left(k_x, k_y\right)\right|_{max} \left(\left| P_j\left(k_x+k_x^{m,n}, k_y+k_y^{m,n}\right)\right|^2 +\delta_1\right)} \Delta _1 ,\nonumber \\
&P_{j+1}\left(k_x, k_y\right) = P_{j}\left(k_x, k_y\right) 
+ \frac{\left| O_j\left(k_x-k_x^{m,n}, k_y-k_y^{m,n}\right)\right| O_j^*\left(k_x-k_x^{m,n}, k_y-k_y^{m,n}\right) } {\left| O_j\left(k_x, k_y\right)\right|_{max} \left(\left| O_j\left(k_x-k_x^{m,n}, k_y-k_y^{m,n}\right)\right|^2 +\delta_2\right)} \Delta _2 ,
\label{eq10}
\end{align}
where $\delta_1$ and $\delta_2$ are two regularization constants used to ensure numerical stability, which are set as $\delta_1=1$, $\delta_2=1000$ in this work, and $\Delta_1$ and $\Delta_2$ are defined as
\begin{align}
&\Delta_1 = \varPhi\left(k_x+k_x^{m,n}, k_y+k_y^{m,n}\right)-O_{j}\left(k_x, k_y\right)P_j\left(k_x+k_x^{m,n}, k_y+k_y^{m,n}\right),\nonumber \\
&\Delta_2 = \varPhi\left(k_x, k_y\right)-O_{j}\left(k_x-k_x^{m,n}, k_y-k_y^{m,n}\right)P_j\left(k_x, k_y\right).
\end{align}

Step 5. Repeat steps 2-4 for all of the LEDs. The LED updating range is $S_1=\left\lbrace\left(m,n\right) \mid m=-(R_1+1)/2,...(R_1+1)/2,n=-(R_1+1)/2,...(R_1+1)/2\right\rbrace$, where $R_1$ is the number of the LEDs in each side of the LED array.

Step 6. Repeat steps 2-5 until the algorithm converges.

In order to illustrate the mcFPM better, we first introduce the conventional SA method~\cite{maiden2012annealing,yeh2015experimental}. 
In the conventional SA method, it assumes that each LED has an independent shift. After step 2 of the above FPM algorithm, the SA module is added to search the deviation of the illumination wave vector $\Delta k_x^{m,n}$, $\Delta k_y^{m,n}$, and the corresponding pupil function with the cost function is defined as
\begin{equation}
 E_1 = \min_{\Delta k_x^{m,n}, \Delta k_y^{m,n}} \sum_{x,y} \left| I_{captured}^{m,n}\left(x, y\right) - \left|\psi_j^{m,n}\left(x, y, \Delta k_x^{m,n}, \Delta k_y^{m,n} \right)\right| ^2 \right| ^2,
\label{eq12}
\end{equation}
where $\psi_j^{m,n}(x, y, \Delta k_x^{m,n}, \Delta k_y^{m,n} )$ is the calculated complex field of the LR image according to Eq.~(\ref{eq8}). During the SA process, the updated wave-vectors are
\begin{align}
&k_x^{m,n} = k_x^{m,n}+ \Delta k_x^{m,n} ,\nonumber \\
&k_y^{m,n} = k_y^{m,n}+\Delta k_y^{m,n} .
\label{eq13}
\end{align}
Because the conventional SA method assumes that the shifting value of each LED is independent, it does not provide any constraint on the LEDs' positions. After several iterations in the SA, the LED position coordinate may become disordered in the algorithm, especially in the edge regions of the LED array~\cite{yeh2015experimental}. Besides, for each sub iterations, the SA process is used for every LED position correction, which is heavily time consuming.

In the proposed mcFPM, rather than correcting the wave-vector $(k_x^{m,n}, k_y^{m,n})$ of each LED in the Fourier domain, we directly correct the global shift of the LED array. 

The global shift of the LED array in the mcFPM has been defined in Sec. 2.3. According to Eq.~(\ref{eq4}), the position of each LED with the global shift is $(x_{m,n}, y_{m,n}) = (md + \Delta x, nd + \Delta y)$. Considering the distance between two adjacent LEDs is $d$, the ranges of the global shifts $\Delta x$ and $\Delta y$ are set to $\left[ -d, d\right]$. 
As Fig.~\ref{fig4} shows, in the mcFPM, the initial values of $\Delta x$ and $\Delta y$ are \SI{0}{\milli\meter} respectively. According to Eq.~(\ref{eq5}), the updated incident wave vectors of $(k_x^{m,n},k_y^{m,n})$ is
\begin{align}
	&k_x^{m,n} = -\frac{2\pi}{\lambda}\frac{x_o - md - \Delta x}{\sqrt{(x_o - md - \Delta x)^2 + (y_o- nd - \Delta y)^2 + s^2}},\nonumber \\
	&k_y^{m,n} = -\frac{2\pi}{\lambda}\frac{y_o-nd - \Delta y}{\sqrt{(x_o-md - \Delta x)^2 + (y_o-nd - \Delta y)^2 + s^2}}.
	\label{eq14}
\end{align}
To improve the efficiency, only the incident vectors $(k_x^{m,n},k_y^{m,n})$ within the bright field of the objective are calculated. The LED updating range is $S_2=\left\lbrace \left( m,n\right) \mid m=-(R_2+1)/2, ...(R_2+1)/2,n=-(R_2+1)/2, ...(R_2+1)/2\right\rbrace$ during the FPM reconstruction process. The value of $R_2$ is determined by the number of rows or columns of those central LEDs that correspond to the bright field of the objective. The cost function for searching $\Delta x$ and $\Delta y$ is defined as
\begin{equation}
	E_2 = \min_{\Delta x, \Delta y}\sum_{m,n}\sum_{x,y}\left| I_{captured}^{m,n}(x, y) - I_{FPM}^{m,n}(x, y, \Delta x, \Delta y)\right|^2 ,
	\label{eq15}
\end{equation}
where $I_{FPM}^{m,n}(x, y, \Delta x, \Delta y)$ is the corresponding calculated intensity image using the conventional FPM algorithm~(step 1-6) with a global shift. Compared with $E_1$, $E_2$ is the summation of all LEDs within the bright field of the objective. Therefore the time that the mcFPM calls the optimization algorithm is much less than the conventional method.

Different from the FPM reconstruction process of the conventional SA method, only $R_2\times R_2$ central LR images are used in the mcFPM, since these LR images contain mainly information of the sample. The procedure of the FPM reconstruction process in the mcFPM only iterates 5 times~($J=5$). After the fast FPM reconstruction process finishes~($j=J$), the cost function $E_2$ can be calculated according to Eq.~(\ref{eq8}).

To minimize the cost function $E_2$, there are several searching methods, such as SA algorithm and genetic algorithm. In order to compare with the existing methods, we use the SA algorithm to search $\Delta x$ and $\Delta y$ in our mcFPM. This loop continues until the cost function reaches a minimal value. After correcting the global shift of the LED array, all LR images are used to reconstruct the HR images of the sample using the conventional FPM algorithm. In some cases, after correcting the global shift of the LED array using the mcFPM, one iteration of the conventional SA is used to correct the inherent local subtle position error of each LED. Finally, the degradation of the reconstructed HR amplitude and phase caused by the LED misalignment can be eliminated.

\section{Experimental verification}
\subsection{Mechanical alignment of the FPM system}
\begin{figure}[!h]
	\centering
		\begin{subfigure}[b]{0.5\linewidth}
			\centering
			\begin{tikzpicture}
			\scope[nodes={inner sep=0,outer sep=0}]
			\node[anchor=south west]{\includegraphics[width=\linewidth]{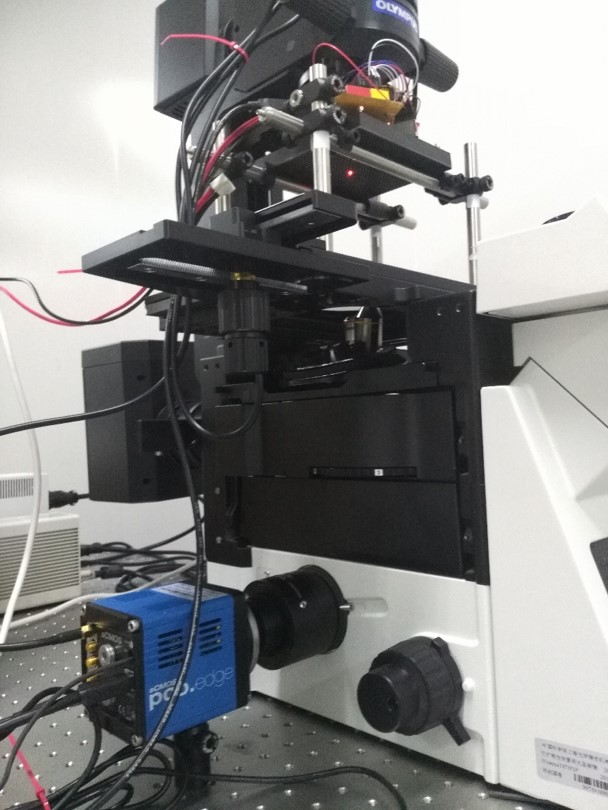}};
			\draw[yellow, line width=0.4mm] (1.9,6.3) rectangle (5.3,8);
			\draw [->,>=stealth, yellow, thick] (5.3,8) -- (6.7,8.5);
			\draw [->,>=stealth, yellow, thick] (5.3,6.3) -- (6.7,5.6);
			\draw[blue, line width=0.4mm] (0.7,0.6) rectangle (3,2.7);
			\draw [->,>=stealth, blue, thick] (3,2.7) -- (6.7,4.6);
			\draw [->,>=stealth, blue, thick] (3,0.6) -- (6.7,0.4);
			\draw[red, line width=0.4mm] (3.7,6.85) rectangle (4,7.1);
			\endscope
			\end{tikzpicture}
			\caption{}
		\end{subfigure}
	\begin{minipage}[b]{0.35\linewidth}
		\begin{subfigure}[b]{\linewidth}
			\centering
			\begin{tikzpicture}
				\scope[nodes={inner sep=0,outer sep=0}]
				\node[anchor=south west]{\includegraphics[width=\linewidth]{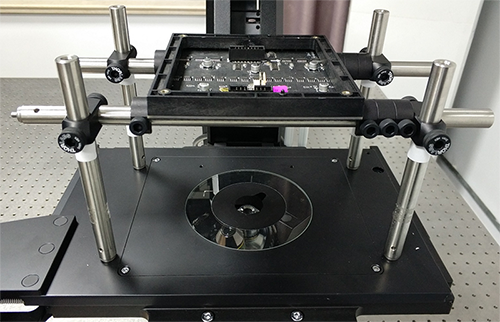}};
				\draw[thick,->,yellow] (1.3,1.05) -- (3.2, 1.05) node[anchor=north west] {$\textcolor{yellow}{x}$};
				\draw[thick,->,yellow] (1.9, 0.5) -- (2.64, 1.5) node[anchor=north west] {$\textcolor{yellow}{y}$};
				\draw[thick,->,yellow] (2.31, 1.05) -- (2.31, 1.6) node[anchor=north west] {$\textcolor{yellow}{z}$};
				\draw (2.4, 0.9) node {$\textcolor{yellow}{0}$}; 
				\endscope
			\end{tikzpicture}
			\caption{}
		\end{subfigure}
		
		\begin{subfigure}[b]{\linewidth}
			\centering
			\includegraphics[width=\linewidth]{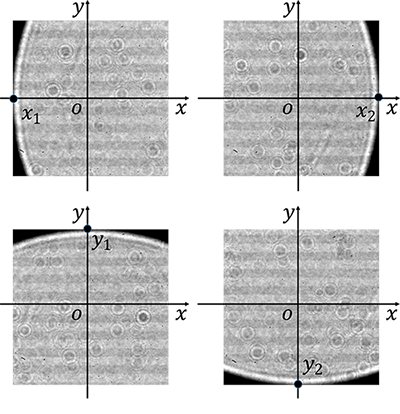}
			\caption{}
		\end{subfigure}
	\end{minipage}
	\caption{(a) is the experimental setup. (b) shows how we install the LED array~(the yellow box of (a)) on the microscope. (c) shows a method to adjust the central LED~(the red box of (a)) along the optical axis~(the z-axis in (b)).}
	\label{fig5}
\end{figure}
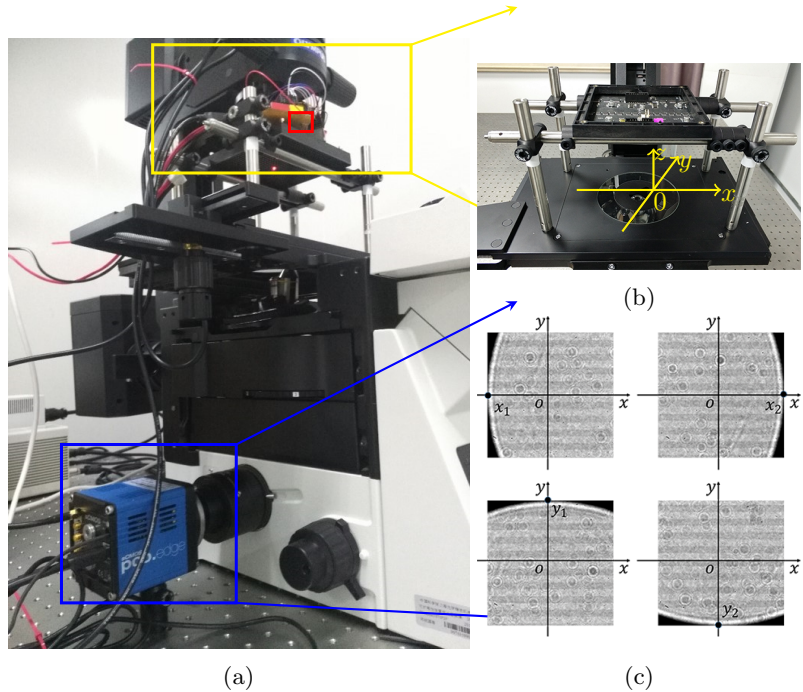

The system setup in this work is shown in Fig.~\ref{fig5}(a). We built an FPM system by replacing the light source of an Olympus IX 73 inverted microscope with a programmable LED array ($32\times32$ LEDs, \SI{4}{\milli\meter} spacing) controlled by an Arduino. The LEDs have a central wavelength of \SI{629}{\nano\meter} and a bandwidth of \SI{20}{\nano\meter}. All samples were imaged with a 4$\times$ 0.1 NA objective and a scientific complementary metal oxide semiconductor~(sCMOS) camera (PCO. edge 4.2). The pixel size of the sCMOS camera is \SI{6.5}{\micro\meter}. The distance between the sample and the LED array is \SI{113.5}{\milli\meter}. In the experiment, the central $17\times17$ LEDs were switched on sequentially to capture 289 LR intensity images, in which only the LR images of the central $5\times5$ LEDs are within the bright field range of the objective. This means that $R_1=17$ and $R_2=5$ in this work. 

Coarse adjustment for the position of the LED array was first performed before the image acquisition. After capturing all the LR images, the proposed algorithm was applied to realize fine correction in the image reconstruction process. As Fig.~\ref{fig5}(b) shows, in order to avoid the position error caused by the rotation of the LED array, we used a set of rods and buckles to fix it on the microscope stage. Then, a level instrument was used to make the sample plane and the LED array plane parallel. Figure~\ref{fig5}(c) shows the principle of the coarse adjustment to make the central LED on the optical axis~(the z-axis in (b)). First, we turned on the central LED~(the red box of Fig.~\ref{fig5}(a)) without a sample placed on the stage. Then we adjusted the stage along the x-axis and y-axis to find four critical positions $x_1, x_2, y_1, y_2$ respectively. The light spot is tangential to the image borders of the camera~(the blue box of Fig.~\ref{fig5}(a)) at these positions. Finally, we moved the stage to the coordinate of $((x_1 + x_2)/2, (y_1 + y_2)/2)$. After this process, the rotation effect is almost eliminated.

\begin{figure}[!h]
	\centering
	\captionsetup[subfigure]{justification=centering}
		\begin{subfigure}[b]{0.4\linewidth}
			\centering
			\begin{tikzpicture}
			    \scope[nodes={inner sep=0, outer sep= 1pt}]
				\node[anchor=south west]{
				\includegraphics[width=\linewidth]{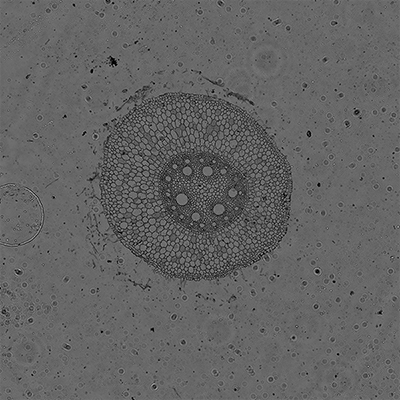}};
				\draw[yellow](0.5,4.5) rectangle (0.85,4.85) node [anchor=west]{b};
				\draw[yellow](1.,3.95) rectangle (1.35,4.3) node [anchor=west]{c};
				\draw[yellow](1.55,3.45) rectangle (1.9,3.8) node [anchor=west]{d};
				\draw[yellow](2.05,2.9) rectangle (2.4,3.25) node [anchor=west]{e};
				\draw[yellow](2.6,2.4) rectangle (2.95,2.75) node [anchor=west]{f};
				\draw[yellow](3.4,1.6) rectangle (3.75,1.95) node [anchor=west]{g};
				\draw[yellow, line width=1mm](0.4,0.2) --node [color=yellow,pos=0.4,above,sloped]{\SI{500}{\micro\meter}} (1.4,0.2) ;
				\endscope
			\end{tikzpicture}
			\caption{}
		\end{subfigure}
	\begin{minipage}[b]{0.58\linewidth}
		\centering
		\begin{subfigure}[b]{0.3\linewidth}
			\begin{tikzpicture}
			\scope[nodes={inner sep=0,outer sep=0}]
			\node[anchor=south west]{\includegraphics[width=\linewidth]{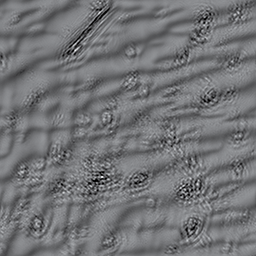}};
			\draw[yellow, line width=1mm](0.3,0.2) --node [color=yellow,pos=0.5,above,sloped]{\SI{40}{\micro\meter}} (0.8257,0.2) ;
			\endscope
			\end{tikzpicture}
			\caption{}
		\end{subfigure}
		\begin{subfigure}[b]{0.3\linewidth}
			\includegraphics[width=\linewidth]{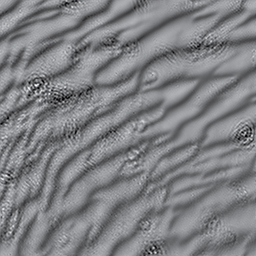}
			\caption{}
		\end{subfigure}
		\begin{subfigure}[b]{0.3\linewidth}
			\includegraphics[width=\linewidth]{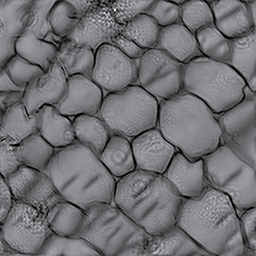}
			\caption{}
		\end{subfigure}
		
		\begin{subfigure}[b]{0.3\linewidth}
			\includegraphics[width=\linewidth]{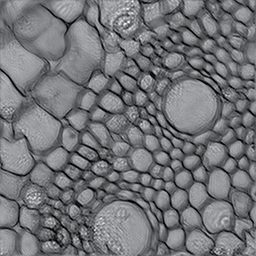}
			\caption{}
		\end{subfigure}
		\begin{subfigure}[b]{0.3\linewidth}
			\includegraphics[width=\linewidth]{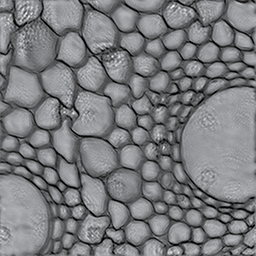}
			\caption{}
		\end{subfigure}
		\begin{subfigure}[b]{0.3\linewidth}
			\includegraphics[width=\linewidth]{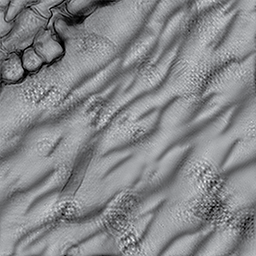}
			\caption{}
		\end{subfigure}
	\end{minipage}
	\caption{\label{fig6} The reconstructions influenced by the global shift of the LED array in the FPM. (a) is one of the original captured images. (b)-(g) are the reconstructed HR intensity images corresponding to the different segments of (a) respectively.}
\end{figure}
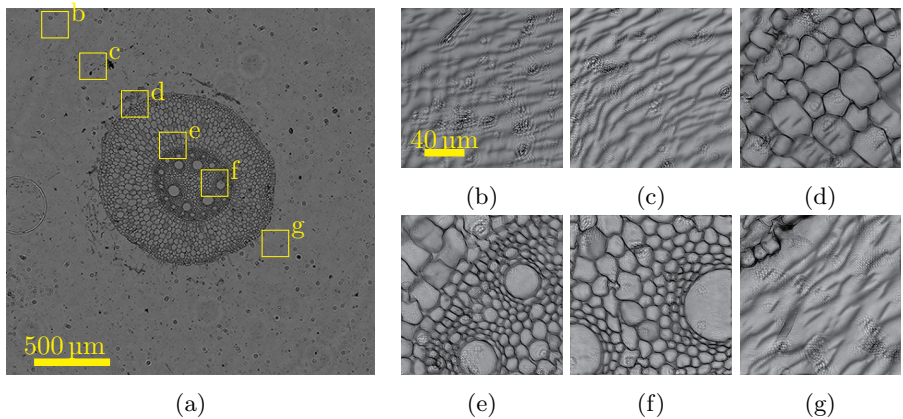

\subsection{The reconstructions influenced by the global shift of the LED array} 
To observe the influence of the global shift in the FPM, we reconstructed the segments of the sample in different regions within the FOV. The FPM reconstruction algorithm we used in this study is adapted from the open source code published in Ref.~\cite{tian2014multiplexed}. Figure~\ref{fig6} shows the reconstructed results using the FPM algorithm without any position correction. Figure~\ref{fig6}(a) is the captured original image of a young plant root sample. Figure~\ref{fig6}(b)-\ref{fig6}(g) are the reconstructed HR intensity images in the different regions of Fig.~\ref{fig6}(a). From the reconstructed results, we can find that the central regions ((e) and (f)) are much better than that of the edge regions ((b) and (c)). In the edge regions, obvious stripes are clearly observable. Based on the phenomenon shown by Fig. 3, we speculate that different global shifts occur at different regions of the FOV. In our coarse adjustment, we carefully aligned the LED array and the camera so that the central LED was imaged to the center of the camera, therefore the influence of global shift near the center regions of the reconstructed image is small. But for the edge regions, aberration of the objective may also affect the reconstruction. This global shift can not be corrected by the FPM algorithm with a correct pupil function.

\subsection{Performance verifications of the proposed mcFPM}

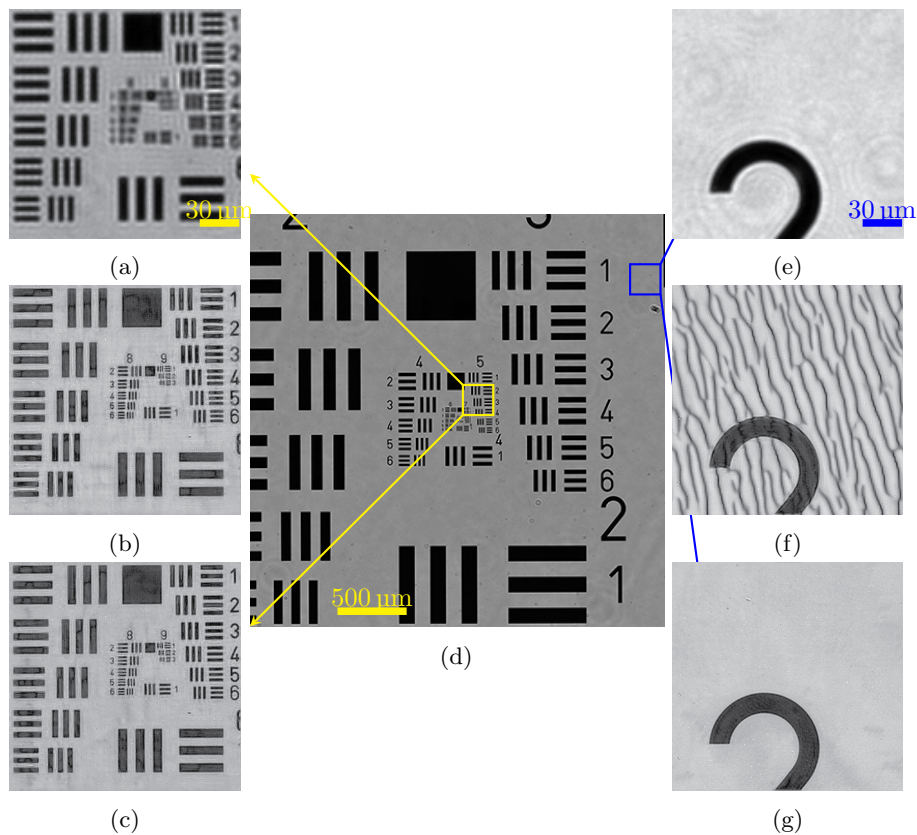
\begin{figure}[htbp]
	\centering
	\captionsetup[subfigure]{justification=centering}
	\begin{subfigure}[b]{0.25\linewidth}
	\begin{minipage}[c]{1\textwidth}
		\centering
		\begin{tikzpicture}
		\scope[nodes={inner sep=0,outer sep=0}]
		\node[anchor=south west]{\includegraphics[width=1\linewidth]{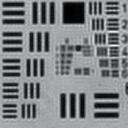}};
		\draw[yellow, line width=1mm](2.5,0.2) --node [color=yellow,pos=0.5,above,sloped]{\SI{30}{\micro\meter}} (3.02,0.2) ;
		\endscope
		\end{tikzpicture}
		\caption{}

		\includegraphics[width=1\linewidth]{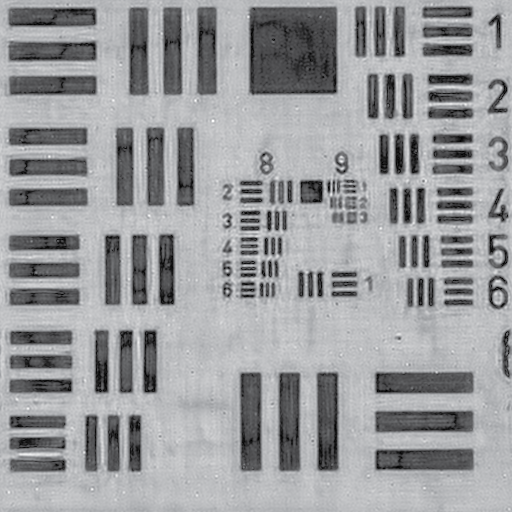}
        \caption{}

		\includegraphics[width=1\linewidth]{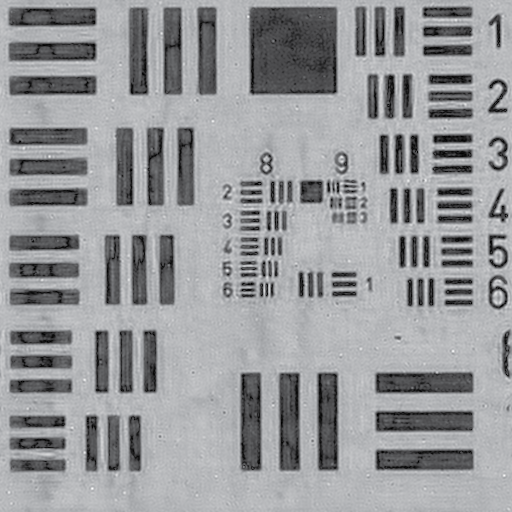}
		\caption{}
	\end{minipage}
	\end{subfigure}	
	\begin{subfigure}[b]{0.45\linewidth}
	\begin{minipage}[c]{1\textwidth}
		\centering
		\begin{tikzpicture}
			\scope[nodes={inner sep=0,outer sep=0}]
			\node[anchor=south west]{\includegraphics[width=1\linewidth]{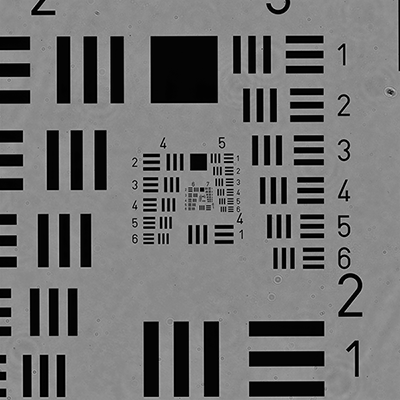}};
			\draw[blue, thick] (5.,4.4) rectangle (5.4,4.8);
			\draw [->,>=stealth, blue, thick] (5.4,4.8) -- (6,6);
			\draw [->,>=stealth, blue, thick] (5.4,4.4) -- (6,0);
			\draw[yellow, thick] (2.8,2.8) rectangle (3.2,3.2);
			\draw [->,>=stealth, yellow, thick] (2.8,3.2) -- (0,6);
			\draw [->,>=stealth, yellow, thick] (2.8,2.8) -- (0,0);
			\draw[yellow, line width=1mm](1.15,0.2) --node [color=yellow,pos=0.5,above,sloped]{\SI{500}{\micro\meter}} (2.06,0.2) ;
			\endscope
			\end{tikzpicture}
		\caption{}
	\end{minipage}
	\end{subfigure}	
	\begin{subfigure}[b]{0.25\linewidth}
	\begin{minipage}[c]{1\textwidth}
		\centering
		\begin{tikzpicture}
		\scope[nodes={inner sep=0,outer sep=0}]
		\node[anchor=south west]{\includegraphics[width=1\linewidth]{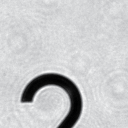}};
		\draw[blue, line width=1mm](2.5,0.2) --node [color=blue,pos=0.5,above,sloped]{\SI{30}{\micro\meter}} (3.02,0.2) ;
		\endscope
		\end{tikzpicture}
		\caption{}
        
        \includegraphics[width=1\linewidth]{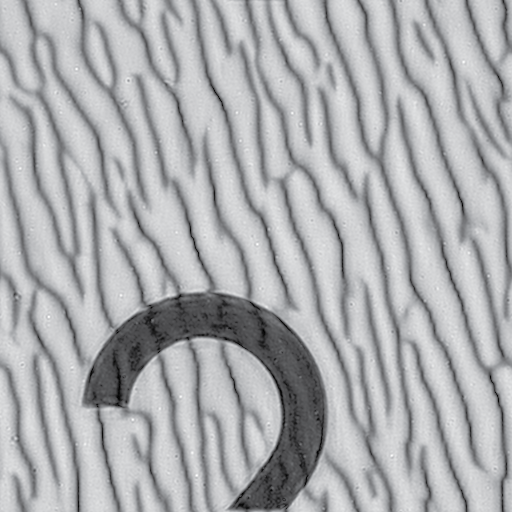}
		\caption{}

		\includegraphics[width=1\linewidth]{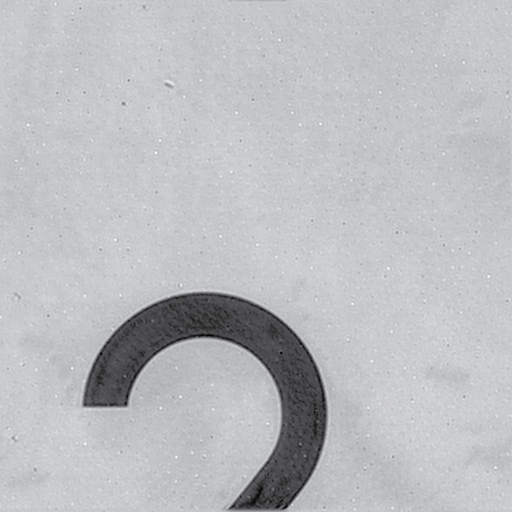}
		\caption{}
	\end{minipage}
	\end{subfigure}
	\caption{\label{fig7} Experimental results of two segments of a USAF resolution target reconstructed with the conventional FPM and the mcFPM. (a) and (e) are the enlargement of the yellow box in (d) the original captured image, (b) and (f) are the reconstructed HR intensity images using the conventional FPM without the position correction, (c) and (g) are the reconstructed HR intensity images using the mcFPM.}
\end{figure}

To verify the feasibility of the proposed method, we compare the reconstructed results of different samples with and without LED position correction. Figure~\ref{fig7} shows the recovered images of two segments of a USAF resolution target. Figure~\ref{fig7}(d) is the original captured image, fig.~\ref{fig7}(a) and ~\ref{fig7}(e) show the enlargement of two different parts~($128\times128$ pixels) of it. Figure~\ref{fig7}(b) and ~\ref{fig7}(f) show the reconstructed HR intensity images using the FPM algorithm without any position correction. Figure~\ref{fig7}(c) and ~\ref{fig7}(g) show the reconstructed results using the mcFPM. Compared with Fig.~\ref{fig7}(b) and \ref{fig7}(f), Fig. ~\ref{fig7}(c) and \ref{fig7}(g) have much higher visible image quality.

\begin{figure}[!h]
	\centering
	\captionsetup[subfigure]{justification=centering}
	\begin{subfigure}[b]{0.45\linewidth}
		\centering
		\begin{tikzpicture}
		\scope[nodes={inner sep=0,outer sep=0}]
		\node[anchor=south west]{\includegraphics[width=1\linewidth]{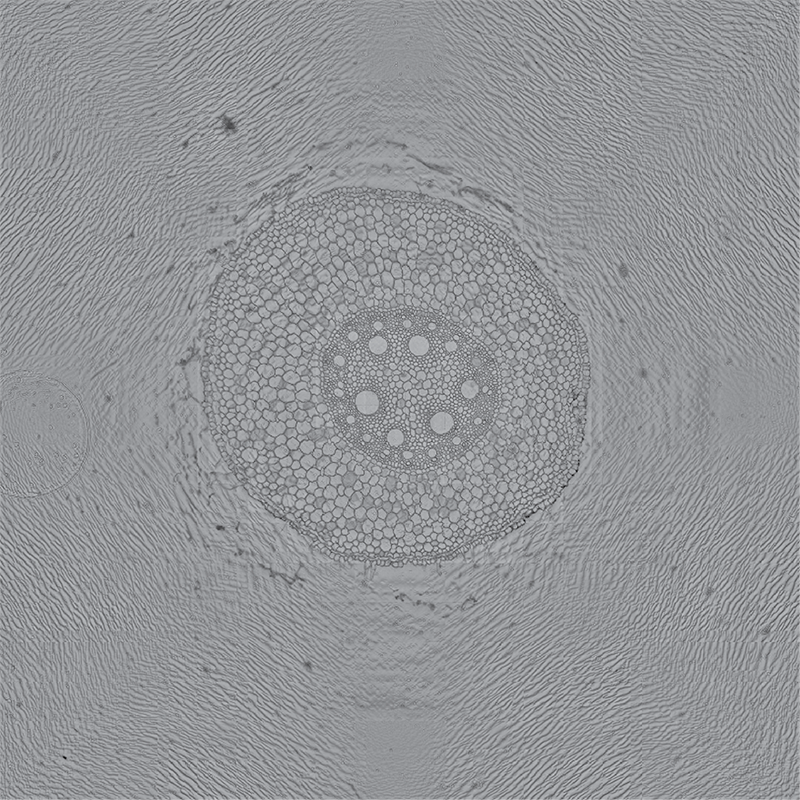}};
		\draw[yellow, line width=1mm](0.4,0.2) --node [color=yellow,pos=0.5,above,sloped]{\SI{500}{\micro\meter}} (1.5367,0.2) ;
		\endscope
		\end{tikzpicture}
		\caption{}
	\end{subfigure}	
	\begin{subfigure}[b]{0.45\linewidth}
		\centering
		\begin{tikzpicture}
		\scope[nodes={inner sep=0,outer sep=0}]
		\node[anchor=south west]{\includegraphics[width=1\linewidth]{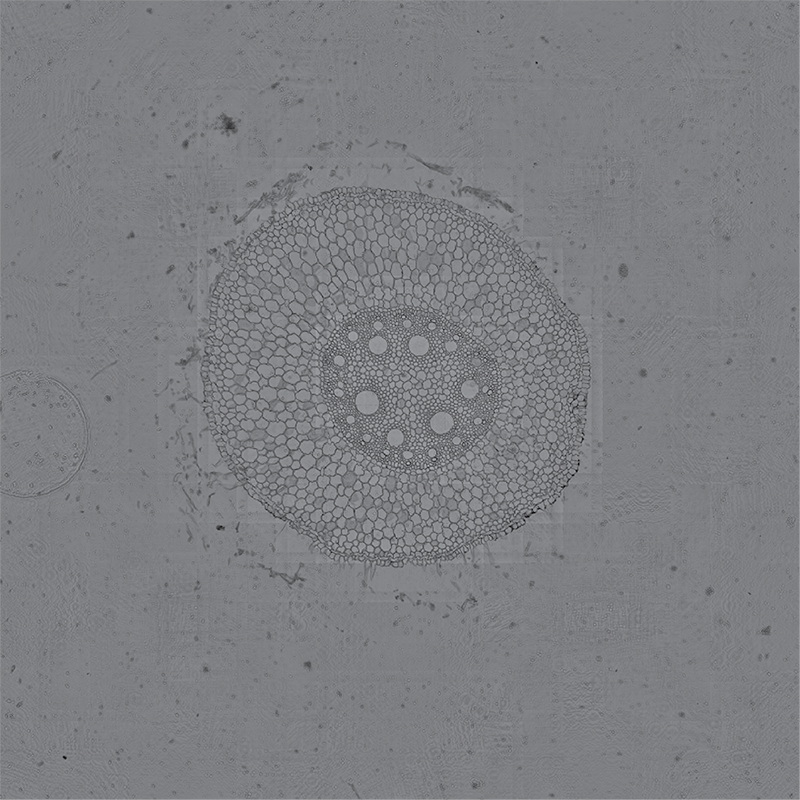}};
		\draw[yellow, line width=1mm](0.4,0.2) --node [color=yellow,pos=0.5,above,sloped]{\SI{500}{\micro\meter}} (1.5367,0.2) ;
		\endscope
		\end{tikzpicture}
		\caption{}
	\end{subfigure}
	\caption{\label{fig8} The reconstructed wide FOV and HR images using (a) the conventional FPM without any position correction and (b) the mcFPM.}
\end{figure}

To test the robustness of the mcFPM in different segments of the FOV, we recovered an HR image with full FOV of a young plant root sample. The full FOV images~($8112 \times 8112$ pixels) were reconstructed as Fig.~\ref{fig8} shows. Figure~\ref{fig8}(a) shows the reconstructed images using the conventional FPM algorithm without any position corrections. The image quality in the center FOV is good while the stripes become obvious in the edge regions. Therefore, the ability of gigapixel imaging with the FPM is seriously decreased because of the regionally dependent global shift of the LED array. Figure~\ref{fig8}(b) shows the reconstructed result with our proposed mcFPM. Compared with the reconstructed results without position correction, we can find that the stripes in the edge regions of the FOV are significantly eliminated, and the image quality is improved greatly as well. This clearly demonstrates the robustness of the proposed mcFPM.

\begin{figure}[!h]
	\centering
	\captionsetup[subfigure]{justification=centering}
	\begin{subfigure}[b]{0.49\columnwidth}
		\centering
		\includegraphics[width=0.48\columnwidth]{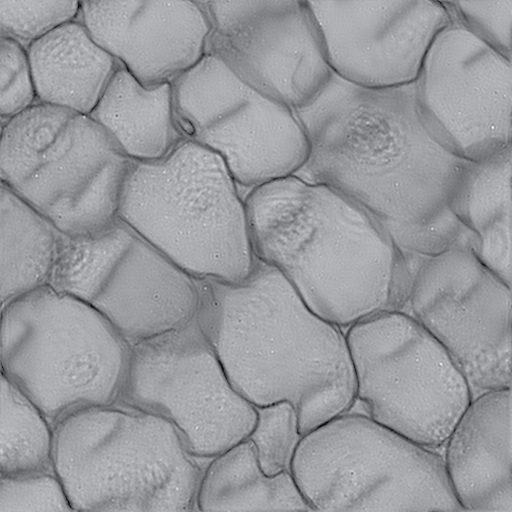}
		\includegraphics[width=0.48\columnwidth]{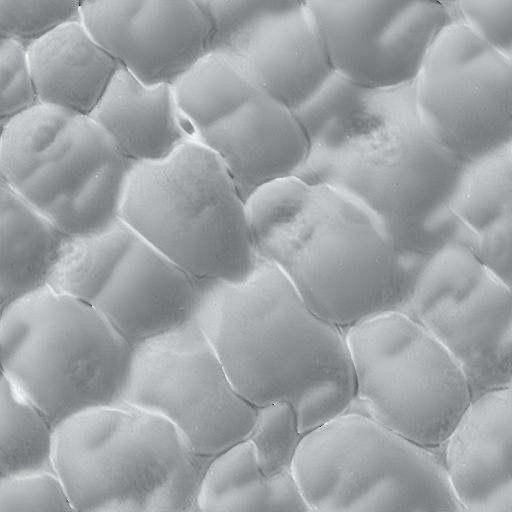}
		\caption{}
	\end{subfigure}	
	\begin{subfigure}[b]{0.49\columnwidth}
		\centering
		\includegraphics[width=0.48\columnwidth]{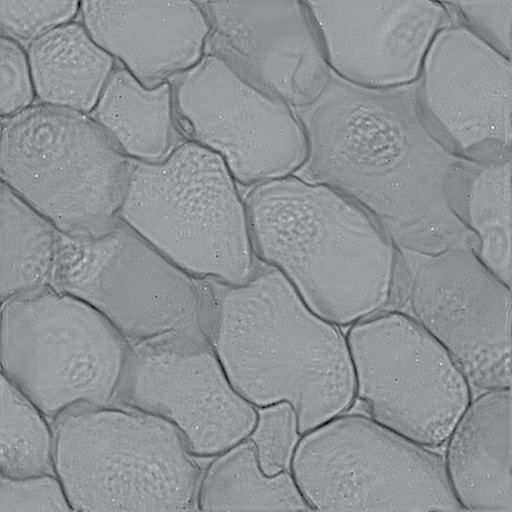}
		\includegraphics[width=0.48\columnwidth]{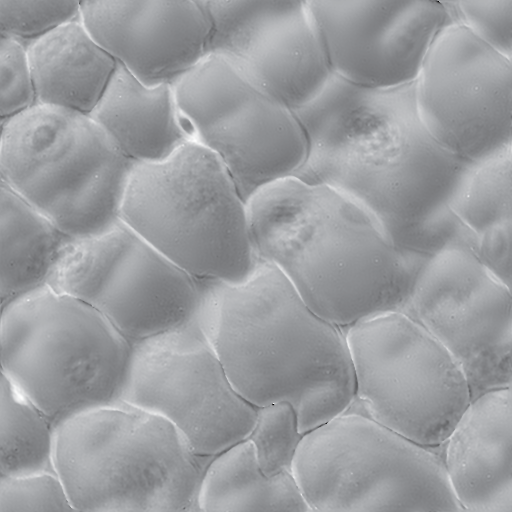}
		\caption{}
	\end{subfigure}
	
	\begin{subfigure}[b]{0.49\columnwidth}
		\centering
		\includegraphics[width=0.48\columnwidth]{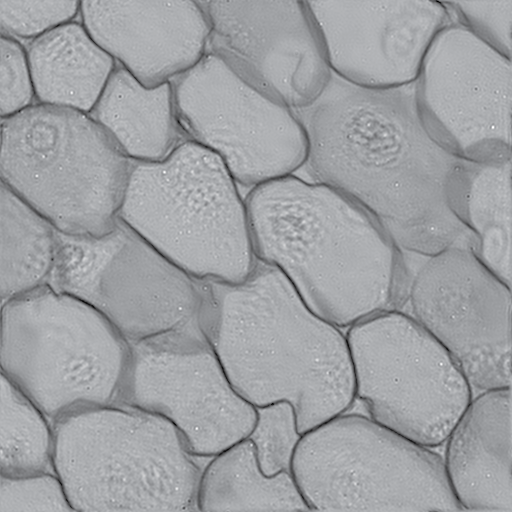}
		\includegraphics[width=0.48\columnwidth]{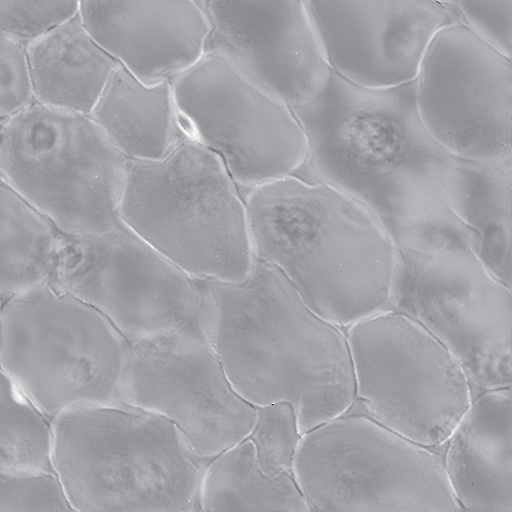}
		\caption{}
	\end{subfigure}	
	\begin{subfigure}[b]{0.49\columnwidth}
		\centering
		\includegraphics[width=0.48\columnwidth]{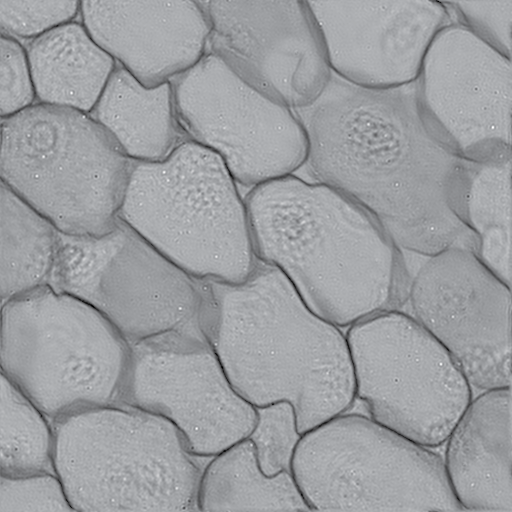}
		\includegraphics[width=0.48\columnwidth]{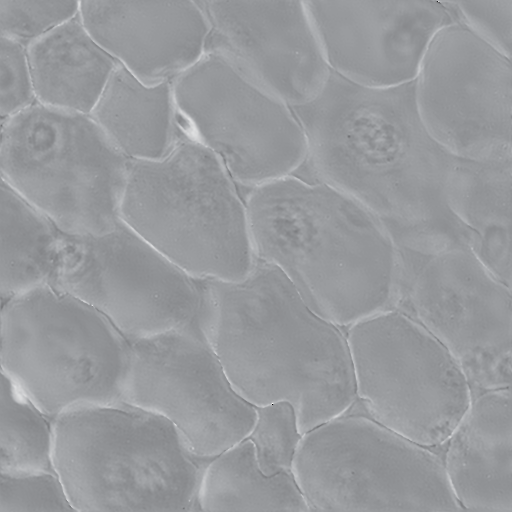}
		\caption{}
	\end{subfigure}
	\caption{\label{fig9} The reconstructed results using different methods: (a) FPM without position correction, (b) Conventional SA method, (c) pcFPM, (d) mcFPM.}
\end{figure}

Furthermore, we compare the proposed mcFPM with the other existing techniques~(conventional SA and pcFPM). The experimental LR images in the edge regions of the sample were chosen as the test data. The pixel sizes of the LR images and the reconstructed HR images are $128 \times 128$ and $512 \times 512$ respectively. The codes were run with MATLAB R2015b on a Windows 10 Enterprise Edition operation system~(Inter i7-6700 CPU @3.40Ghz, 8 GB DDR4 memory). The SA algorithm terminates when the average change of the cost function is less than $10^{-3}$, or the number of iterations exceeds 100. Figure~\ref{fig9} shows the reconstructed results using different methods. The left images of each sub figure are the reconstructed amplitude profiles and the right ones are the phase profiles. Figure~\ref{fig9}(a) shows stripes on both of the amplitude and phase. As shown from Fig.~\ref{fig9}(b), conventional SA method can reconstruct a good HR amplitude, but the stripes in the phase still exist. Besides, it takes \SI{246}{\second} for one segment. The reason is that it calls SA algorithm once for each LED. As Fig.~\ref{fig9}(c) shows, the pcFPM removes the stripes in both the amplitude and phase, however, it takes \SI{728}{\second}. Figure~\ref{fig9}(d) shows the reconstructed results using the mcFPM. Compared with the previous methods, the mcFPM takes the least time to run, which is only \SI{40}{\second}. On the whole, the conventional SA method is time consuming, and breaks the physical constraint of the LED array. The pcFPM takes the physical constraint into account, but it takes more time because of the additional optimization process. Our method directly correct the global shift of the LED array, which is the most efficient among all the testing algorithms. 

\section{Conclusion}
In the FPM, the misalignment limits its capability to realize gigapixel imaging. By analyzing the experimental data, we found that different regions of the FOV have different global LED position shifts. This produces observable stripes in the reconstructed HR images. To eliminate the global shift of the LED array, we have proposed the mcFPM algorithm. Rather than correcting the shift errors of the pupil function in the Fourier domain, we introduced a global position misalignment model of the LED array with two factors, and then directly corrected the global shift of the LED array. Experimental results have shown that the mcFPM performs robustly in different regions of the FOV. The experiments have shown the mcFPM is more efficiently than all of the state-of-the-art techniques.

\section*{Funding}
National Natural Science Foundation of China~(NSFC)~(61327902, 61705241); Natural Science Foundation of Shanghai~(NSFS)~(17ZR1433800); and the Chinese Academy of Sciences (QYZDB-SSW-JSC002).



\begin{thebibliography}{10}

\bibitem{bian2013adaptive}
Zichao Bian, Siyuan Dong, and Guoan Zheng.
\newblock Adaptive system correction for robust {Fourier} ptychographic
  imaging.
\newblock {\em Optics Express}, 21(26):32400--32410, 2013.

\bibitem{Chen_2014_JOSK}
Ni~Chen, Jiwoon Yeom, Keehoon Hong, Gang Li, and Byoungho Lee$^*$.
\newblock Fast converging algorithm for wavefront reconstruction based on a
  sequence of diffracted intensity images.
\newblock {\em Journal of the Optical Society of Korea}, 18(3):217--224, 2014.

\bibitem{chung2015counting}
Jaebum Chung, Xiaoze Ou, Rajan~P Kulkarni, and Changhuei Yang.
\newblock Counting white blood cells from a blood smear using {Fourier}
  ptychographic microscopy.
\newblock {\em PLOS One}, 10(7):e0133489, 2015.

\bibitem{faulkner2004movable}
H.~M.~L. Faulkner and J.~M. Rodenburg.
\newblock Movable aperture lensless transmission microscopy: a novel phase
  retrieval algorithm.
\newblock {\em Physical Review Letters}, 93(2):023903, 2004.

\bibitem{fienup1982phase}
James~R Fienup.
\newblock Phase retrieval algorithms: a comparison.
\newblock {\em Applied Optics}, 21(15):2758--2769, 1982.

\bibitem{horstmeyer2015digital}
Roarke Horstmeyer, Xiaoze Ou, Guoan Zheng, Phil Willems, and Changhuei Yang.
\newblock Digital pathology with {Fourier} ptychography.
\newblock {\em Computerized Medical Imaging and Graphics}, 42:38--43, 2015.

\bibitem{maiden2012annealing}
A.M. Maiden, M.J. Humphry, M.C. Sarahan, B.~Kraus, and J.M. Rodenburg.
\newblock An annealing algorithm to correct positioning errors in ptychography.
\newblock {\em Ultramicroscopy}, 120:64--72, 2012.

\bibitem{maiden2009improved}
Andrew~M. Maiden and John~M. Rodenburg.
\newblock An improved ptychographical phase retrieval algorithm for diffractive
  imaging.
\newblock {\em Ultramicroscopy}, 109(10):1256--1262, 2009.

\bibitem{ou2013quantitative}
Xiaoze Ou, Roarke Horstmeyer, Changhuei Yang, and Guoan Zheng.
\newblock Quantitative phase imaging via {Fourier} ptychographic microscopy.
\newblock {\em Optics Letters}, 38(22):4845--4848, 2013.

\bibitem{ou2014embedded}
Xiaoze Ou, Guoan Zheng, and Changhuei Yang.
\newblock Embedded pupil function recovery for {Fourier} ptychographic
  microscopy.
\newblock {\em Optics Express}, 22(5):4960--4972, 2014.

\bibitem{rodenburg1992theory}
J.~M. Rodenburg and R.~H.~T. Bates.
\newblock The theory of super-resolution electron microscopy via
  wigner-distribution deconvolution.
\newblock {\em Philosophical Transactions of the Royal Society A},
  339(1655):521--553, 1992.

\bibitem{rodenburg2007hard}
J.~M. Rodenburg, A.~C. Hurst, A.~G. Cullis, B.~R. Dobson, F.~Pfeiffer, O.~Bunk,
  C.~David, K.~Jefimovs, and I.~Johnson.
\newblock Hard-x-ray lensless imaging of extended objects.
\newblock {\em Physical Review Letters}, 98(3):034801, 2007.

\bibitem{sun2016efficient}
Jiasong Sun, Qian Chen, Yuzhen Zhang, and Chao Zuo.
\newblock Efficient positional misalignment correction method for {Fourier}
  ptychographic microscopy.
\newblock {\em Biomedical Optics Express}, 7(4):1336--1350, 2016.

\bibitem{tian2014multiplexed}
Lei Tian, Xiao Li, Kannan Ramchandran, and Laura Waller.
\newblock Multiplexed coded illumination for {Fourier} ptychography with an led
  array microscope.
\newblock {\em Biomedical Optics Express}, 5(7):2376--2389, 2014.

\bibitem{williams2014fourier}
Anthony Williams, Jaebum Chung, Xiaoze Ou, Guoan Zheng, Siddarth Rawal, Zheng
  Ao, Ram Datar, Changhuei Yang, and Richard Cote.
\newblock {Fourier} ptychographic microscopy for filtration-based circulating
  tumor cell enumeration and analysis.
\newblock {\em Journal of Biomedical Optics}, 19(6):066007--066007, 2014.

\bibitem{yeh2015experimental}
Li-Hao Yeh, Jonathan Dong, Jingshan Zhong, Lei Tian, Michael Chen, Gongguo
  Tang, Mahdi Soltanolkotabi, and Laura Waller.
\newblock Experimental robustness of {Fourier} ptychography phase retrieval
  algorithms.
\newblock {\em Optics Express}, 23(26):33214--33240, 2015.

\bibitem{zheng2014breakthroughs}
Guoan Zheng.
\newblock Breakthroughs in photonics 2013: {Fourier} ptychographic imaging.
\newblock {\em IEEE Photonics Journal}, 6(2):1--7, 2014.

\bibitem{zheng2015fourier}
Guoan Zheng.
\newblock {Fourier} ptychographic imaging.
\newblock In {\em Photonics Conference~(IPC), 2015}, pages 20--21. IEEE, 2015.

\bibitem{zheng2013wide}
Guoan Zheng, Roarke Horstmeyer, and Changhuei Yang.
\newblock Wide-field, high-resolution {Fourier} ptychographic microscopy.
\newblock {\em Nature Photonics}, 7(9):739--745, 2013.

\end{thebibliography}

\end{document}